\documentclass[10pt,twocolumn,letterpaper]{article}

\usepackage{cvpr}              

\definecolor{cvprblue}{rgb}{0.21,0.49,0.74}
\usepackage[pagebackref,breaklinks,colorlinks,allcolors=cvprblue]{hyperref}
\usepackage{graphicx}
\usepackage{tabularx}
\usepackage{wrapfig}
\usepackage{multirow}
\usepackage{color}
\usepackage{float}

\def\paperID{6175} 
\def\confName{CVPR}
\def\confYear{2025}

\title{Towards Explicit Geometry-Reflectance  Collaboration for Generalized \\LiDAR Segmentation in Adverse Weather}
\author{Longyu Yang$^1$,~~Ping Hu$^{1}$\thanks{Corresponding author}~,~~Shangbo Yuan$^1$,~~Lu Zhang$^2$,~~Jun Liu$^3$,~~Hengtao Shen$^{1,4}$,~~Xiaofeng Zhu$^1$
\and $^1$UESTC~~~~~$^2$DLUT~~~~~$^3$Lancaster University~~~~~$^4$Tongji University
}

\begin{document}
\maketitle
\begin{abstract}
Existing LiDAR semantic segmentation models often suffer from decreased accuracy when exposed to adverse weather conditions. 
Recent methods addressing this issue focus on enhancing training data through weather simulation or universal augmentation techniques. 
However, few works have studied the negative impacts caused by the heterogeneous domain shifts in the geometric structure and reflectance intensity of point clouds. 
In this paper, we delve into this challenge and address it with a novel Geometry-Reflectance Collaboration (GRC) framework that explicitly separates feature extraction for geometry and reflectance. 
Specifically, GRC employs a dual-branch architecture designed to independently process geometric and reflectance features initially, thereby capitalizing on their distinct characteristic. 
Then, GRC adopts a robust multi-level feature collaboration module to suppress redundant and unreliable information from both branches.
Consequently, without complex simulation or augmentation, our method effectively extracts intrinsic information about the scene while suppressing interference, thus achieving better robustness and generalization in adverse weather conditions.
We demonstrate the effectiveness of GRC through comprehensive experiments on challenging benchmarks, showing that our method outperforms previous approaches and establishes new state-of-the-art results.
\end{abstract}

\section{Introduction}
\label{sec:intro}

LiDAR semantic segmentation is a fundamental task in 3D scene understanding and plays a crucial role in applications like robotics and autonomous driving~\cite{xiao2024survey,guo2020deep}. It works by predicting a semantic label for each point in a LiDAR scan, thus enabling the recognition of objects and environmental features essential for downstream tasks. With the rapid advancement of deep learning techniques, significant progress has been made in this area~\cite{kong2023rethinking,wu2022point,wu2024point,choy20194d,zhou2020cylinder3d,yang2023swin3d,qi2017pointnet}. However, most traditional research is conducted under relatively idealized conditions, utilizing standard datasets for training and testing that often exclude various interference factors encountered in real-world applications, such as adverse weather conditions like fog, rain, snow, etc~\cite{hahner2021fog,bijelic2020seeing,xiao20233d,hahner2022lidar}. Consequently, although current methods perform well in standard environments, they tend to experience significant performance degradation when exposed to more challenging scenarios~\cite{kong2023robo3d,yan2024benchmarking}. This highlights a critical challenge in the task: the lack of robustness and adaptability of LiDAR semantic segmentation models across different scenarios, especially under adverse weather.

\begin{figure}[t]
    \begin{minipage}[t]{0.5\linewidth}
        \centering
        \includegraphics[width=1\textwidth]{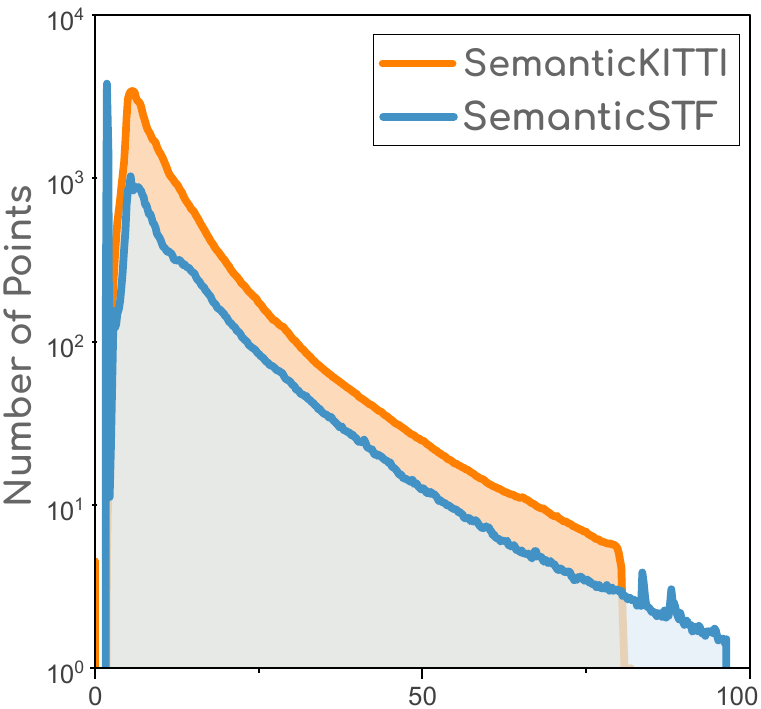}
        \vspace{-0.8cm}
        \caption*{(a) Distance }
        \label{fig:teaser_dis}
    \end{minipage}%
    \begin{minipage}[t]{0.5\linewidth}
        \centering
        \includegraphics[width=1\textwidth]{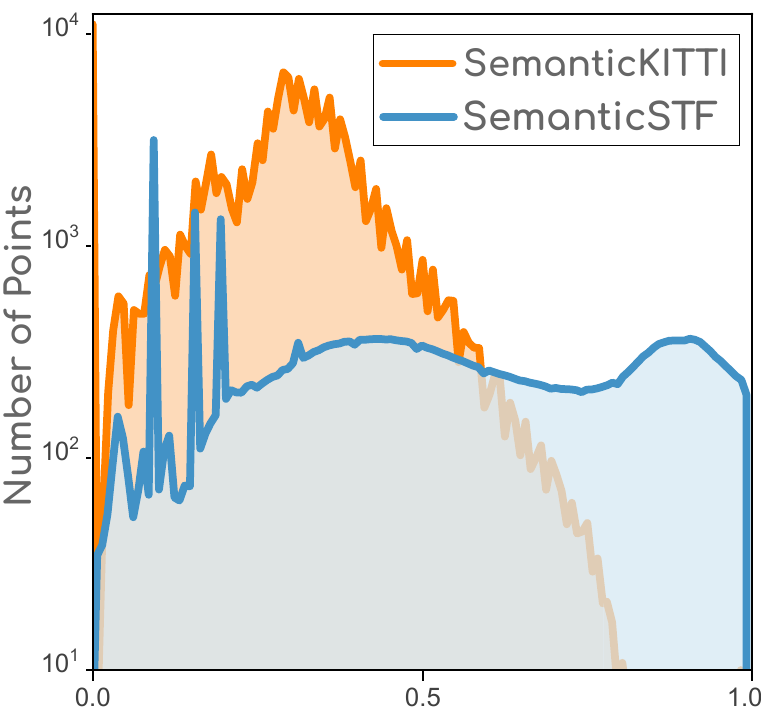}
        \vspace{-0.8cm}
        \caption*{(b) Reflectance}
        \label{fig:teaser_ref}
    \end{minipage}
    \vspace{-0.4cm}
    \caption{Histogram of distance and reflectance intensity for the averaged point cloud in SemanticKITTI~\cite{behley2019semantickitti} and SemanticSTF~\cite{xiao20233d}. Reflectance intensity exhibits a much more pronounced domain shift compared to the geometric layout.}
     \label{fig:teaser}
    \vspace{-0.7cm}
\end{figure}

In addressing the challenge, recent efforts focus on employing weather-simulators~\cite{zhao2024unimix,hahner2022lidar,hahner2021fog,yang2023realistic} and universal augmentation techniques~\cite{xiao20233d,park2024rethinking,kim2023domain,kong2023robo3d,kim2023single,lehner20223d}. Although previous approaches have shown promising results, several limitations may still exist. On the one hand, weather simulation methods synthesize intrinsic characteristics of specific weather conditions but struggle to capture the full spectrum of weather types and severities, leading to limited adaptability when models encounter real-world conditions~\cite{zhao2024unimix}. On the other hand, universal augmentation aims to learn general representations with reduced overfitting to the training data but often produces redundant and futile information, resulting in increased complexity and difficulty in model training~\cite{zhou2022domain,wang2022generalizing}. 
Unlike previous approaches that focus on enhancing the diversity of training data, we present a perspective to achieve robust LiDAR segmentation under adverse weather without relying on complex augmentation or simulation. As plotted in Fig.\ref{fig:teaser}, we observe that adverse weather conditions affect the geometry and reflectance of point clouds in distinct ways. Notably, geometric layout experiences much slighter shifts between normal and adverse conditions, suggesting that reflectance may be a critical factor for performance degradation. We verify the impact of reflectance in Tab.\ref{tab:impact_ref}. As shown, excluding reflectance from input improves generalization performance across four types of adverse weather conditions, demonstrating its negative effects. 

\begin{table}[t]
    \footnotesize
    \begin{tabular}{ccccccc|c}
    \toprule
    Input    & Dense-fog & Light-fog & Rain  & Snow &All \\
    \hline
    MinkNet w/ R  & 29.5 & 26.0 & 28.4 & 21.4 &24.4 \\
    MinkNet w/o R & 32.3 & 37.7 & 35.4 & 32.1 &37.3\\
    \hline
    Ours & 38.1 & 40.1 & 38.1 & 34.1  &42.5\\
    \bottomrule
    \end{tabular}
    \vspace{-0.3cm}
    \caption{Generalized segmentation accuracy of MinkNet~\cite{choy20194d} and the proposed method on the SemanticKITTI$\rightarrow$SemanticSTF task. Excluding reflectance from the input improves MinkNet's performance across various adverse weather conditions.}
    \label{tab:impact_ref}
    \vspace{-0.4cm}
\end{table}


Although discarding reflectance intensity significantly improves generalization capability, this simple strategy can be suboptimal, as corrupted reflectance still contains valuable information beneficial for semantic segmentation. To address this, we propose a novel Geometry-Reflectance Collaboration (GRC) framework designed to effectively extract intrinsic and generalizable features from point clouds. In GRC, we utilize a dual-branch architecture to explicitly separate the processing of geometric and reflectance information, leveraging their distinct and domain-invariant characteristics. Then, a robust multi-level feature collaboration module combines features from both the geometric and reflectance branches, while further suppressing noise and corruption caused by adverse weather. Consequently, our GRC framework effectively reduces mutual interference between geometry and reflectance, allowing for more precise feature extraction and thus achieving robust generalization across adverse conditions without the need for complex data augmentation or simulation, as demonstrated in Tab.~\ref{tab:impact_ref}. Experimental results on challenging benchmark datasets validate the effectiveness and efficiency of our approach. In summary, our contributions are as follows: 
\begin{itemize}
    \item We delve into the challenge in generalized LiDAR segmentation, under heterogeneous domain shifts in geometry and reflectance caused by adverse weather, and propose an effective Geometry-Reflectance Collaboration (GRC) framework to address this issue.
    \item We introduce a Robust Multi-level Feature Collaboration mechanism that effectively harnesses useful information from both geometry and reflectance while mitigating mutual interference.
    \item We conduct extensive experiments on challenging benchmark datasets, demonstrating that the proposed method achieves new state-of-the-art results.
\end{itemize}

\section{Related Work}
\label{sec:related Work}

Point cloud semantic segmentation is an active research area that has witnessed significant advancement with the success of deep learning. Existing methods can be roughly divided into point-based, projection-based, and voxel-based approaches. 
Point-based methods process 3D points directly using multi-layer perceptrons (MLPs) ~\cite{qi2017pointnet,qi2017pointnet++,qian2022pointnext}, kernel point convolutions ~\cite{li2018pointcnn,thomas2019kpconv,wu2019pointconv}, graph neural networks (GNNs) ~\cite{wang2019dynamic,landrieu2018large,huang2022dual}, or transformer architectures ~\cite{zhao2021point,wu2022point,wu2024point,lai2022stratified} to extract features and capture geometric structures.
Projection-based methods transform LiDAR points into 2D images, with rangeview-based techniques ~\cite{cortinhal2020salsanext,milioto2019rangenet++,li2025rapid,ando2023rangevit,sun2024efficient} using spherical projection to create compact, dense, and computationally efficient representations that facilitate semantic segmentation via established 2D convolutional methods and pretrained 2D models, albeit at the cost of potential geometric information loss. 
Voxel-based methods ~\cite{qi2016volumetric,wu20153d,zhu2021cylindrical,lai2023spherical,choy20194d,graham20183d} divide the 3D space into voxel grids~\cite{choy20194d}, cylindrical partitions~\cite{zhu2021cylindrical}, and radial windows~\cite{lai2023spherical}. The sparsity and irregularity of point clouds can lead to redundant computations. It can be addressed by 3D sparse convolutions ~\cite{choy20194d,graham20183d,tang2020searching}, which focus computations on non-empty voxels, thereby enhancing efficiency and reducing computational load and memory usage. 

\begin{figure*}[t]
  \centering
  \includegraphics[width=0.95\linewidth]{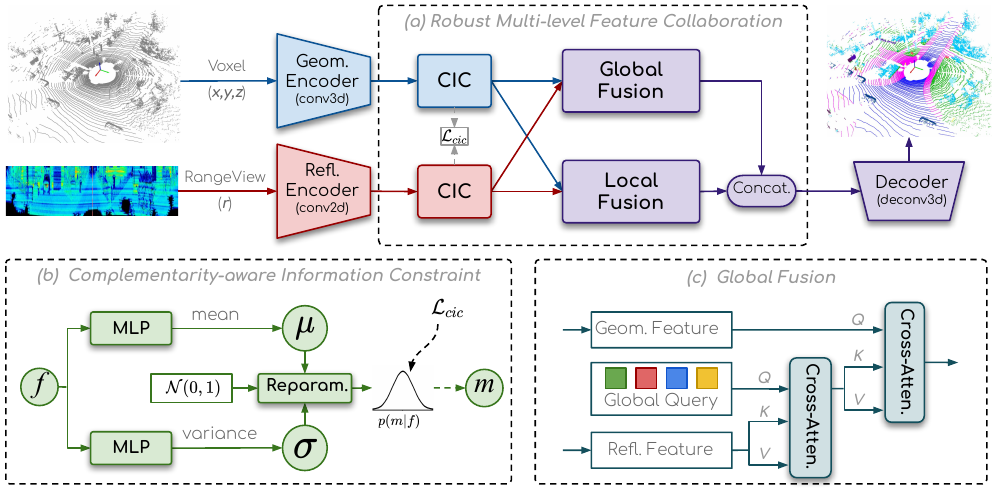}
  \vspace{-0.2cm}
  \caption{ Overview of the Proposed Geometry-Reflectance Collaboration Network (GRCNet). GRCNet begins by independently processing geometric structure and reflectance intensity through separate feature encoders. These features are then integrated using a Robust Multi-level Feature Collaboration (RMFC) module. Within the RMFC, both geometric and reflectance features are augmented by Complementarity-aware Information Constraint (CIC) before being fused at global and local levels. By explicitly separating the feature encoding and then fusing, GRCNet mitigates mutual interference between geometric and reflectance features, enabling the extraction of intrinsic scene features that are robust across various weather conditions.} 
  \vspace{-0.5cm}
  \label{fig:architecture}
\end{figure*}

Despite these advancements, traditional LiDAR segmentation models still face significant challenges in real-world applications, especially when exposed to adverse weather conditions~\cite{hahner2021fog, bijelic2020seeing, hahner2022lidar, yan2024benchmarking, kong2023robo3d}. To mitigate performance degradation, recent efforts have focused on simulating weather-related corruptions in point clouds during training~\cite{zhao2024unimix, hahner2022lidar, hahner2021fog, yang2023realistic}. Although these physically-based strategies are grounded in realistic principles, they are often limited by the specificity of physic model based simulations and may lack adaptability in dynamic and complex environments. To develop more generalizable segmentation models, reducing overfitting through weather-agnostic data augmentation has become a popular approach~\cite{xiao20233d, park2024rethinking, kim2023domain, kong2023robo3d, kim2023single, lehner20223d, he2024domain}. For instance, Xiao~\etal~\cite{xiao20233d} apply geometry style randomization to point clouds, while He~\etal~\cite{he2024domain} introduce augmentation in the feature space. Park~\etal~\cite{park2024rethinking} first evaluate the impact of various universal augmentations and then employ reinforcement learning to predict optimal augmentation strategies. 
Although such universal augmentation based techniques can effectively generate more generalizable features, they often come with increased training costs and complexity~\cite{zhou2022domain, wang2022generalizing}. In contrast to these data-centric approaches, we thoroughly examine adverse weather's heterogeneous impacts on different aspects of point clouds, and accordingly introduce a robust and generalizable segmentation framework without relying on complex data augmentation or simulations.

\section{Methodology}
This section presents the Geometry-Reflectance Collaboration Network (GRCNet), designed for generalized LiDAR semantic segmentation under adverse weather conditions. As illustrated in Fig.\ref{fig:architecture}, GRCNet employs a dual-branch network in the encoding phase to separately process geometric structure and reflectance intensity information, which are then fused to predict semantic labels. We begin by outlining the problem formulation in Sec.\ref{sec:method_fomulation}. Next, in Sec.\ref{sec:method_sfe}, we detail the dual-branch design for separate feature encoding of geometric structure and reflectance intensity. In Sec.\ref{sec:method_umc}, we describe the feature fusion using a Robust Multi-level Feature Collaboration (RMFC) module. Finally, we present the decoder and discuss training details in Sec.~\ref{sec:method_tai}.

\subsection{Problem Formulation}
\label{sec:method_fomulation}
We formulate the task as a domain-generalized LiDAR segmentation problem. The objective is to achieve robust performance on target domains with adverse weather conditions by training exclusively on source domain data under standard conditions. Formally, during training, the model has access to source domain data $\mathcal{S} = \{(P^{s}_{i}, L^{s}_{i})\}_{i=1}^{N^{s}}$, where $N^s$ is the number of scans in the source domain dataset. For the $i$-th source sample $(P^{s}_{i}, L^{s}_{i})$ containing $n$ LiDAR points, $L^{s}_{i} \in \mathbb{R}^{n}$ denotes the point-wise label annotations, and $P^{s}_{i}=\{(x^{s}_{ij},y^{s}_{ij},z^{s}_{ij},r^{s}_{ij})\}^{n}_{j=1}$ represents the $i$-th point cloud with $(x, y, z)$ and $r$ representing coordinates and reflectance intensity of each point respectively. Then, the goal of the task is to learn a semantic segmentation model $f:P\rightarrow L$ with only the source domain data $\mathcal{S}$ to perform well on the target domain $\mathcal{T} = \{P^{t}_{i}\}_{i=1}^{N^{t}}$, which remains unavailable during training.

\subsection{Separated Feature Encoding}
\label{sec:method_sfe}
As discussed in Sec.~\ref{sec:intro}, geometric structure and reflectance intensity capture fundamentally different aspects of LiDAR data: geometric structure represents spatial layout while reflectance intensity relates to surface characteristics and material properties.  These two features undergo distinct degradation when transitioning from standard conditions to adverse weather. Ignoring these inherent differences and treating geometric structure and reflectance intensity as unified inputs can lead to suboptimal feature extraction  when generalizing to challenging weather scenarios. Corruptions or distortions in reflectance can negatively impact the extraction of meaningful geometric features, and vice versa. To address this, we explicitly separate the feature extraction and propose a dual-branch model to independently and specializedly encode geometry and reflectance information. 

\subsubsection{Encoding Geometric Information}
The voxel-based approach demonstrates robustness in dealing with variations in geometric information. The voxelization process discretizes continuous 3D spatial space into a regular grid of voxels, which reduces the sensitivity to noise and distortion. Moreover, operations like Conv3D on voxels can further enhance robustness due to the feature smoothing effect and multi-scale context aggregation. Inspired by this, we exploit voxel-based deep networks for generalizable geometric information encoding.
Formally, we exclude the reflectance intensity $r$ from a input point cloud $P$, resulting in $P_{geo} = [\textbf{x}, \textbf{y}, \textbf{z}] \in \mathbb{R}^{n \times 3}$. Then, $P_{geo}$ is voxelized and further processed by a geometric encoder,
\begin{align}
    F_{geo}=\mathbf{E}_{geo}(P_{geo})
\end{align}
where $F_{geo}\in \mathbb{R}^{n \times c}$ is the resulting geometric feature in the form of sparse representation,  and $E_{geo}$ is the encoder network composed of a series of sparse Conv3D blocks~\cite{choy20194d} that gradually aggregates spatial resolutions and geometric context.

\begin{figure}
    \centering
    \includegraphics[width=\linewidth]{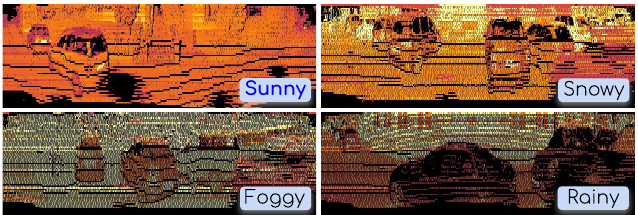}
    \vspace{-0.7cm}
    \caption{Visualization of reflectance intensity under different weather conditions, with intensity values represented in the same color scale.}  
    \vspace{-0.5cm}
    \label{fig:ref_vis}
\end{figure}

\subsubsection{Encoding Reflectance Information} 
Compared to geometric structure, reflectance intensity is typically more affected by adverse weather, as adverse weather can significantly alter or attenuate intensity values through scattering or absorption of the LiDAR signal. This sensitivity to environmental factors causes drastic changes in the distribution of reflectance intensity, hindering the model’s ability to generalize effectively. Nevertheless, despite these substantial variations, reflectance intensity still conveys valuable appearance information that aids in semantic segmentation. As shown in Fig.~\ref{fig:ref_vis}, even with intensity fluctuations and noise, in the dense 2D projection, appearance contours remain distinguishable due to relative intensity differences between objects.

Range view is a natural way to remove 3D layout and better retain 2D appearance in reflectance. To capture weather-invariant semantic information from reflectance intensity, we apply spherical projection~\cite{wu2018squeezeseg} to transform 3D point cloud into 2D range-view image $P_{ref}\in\mathbb{R}^{H\times W}$, where each point's reflectance intensity is retained as input feature for the corresponding pixel. Reflectance features are then extracted as follows,
\begin{align}
    F_{ref}=E_{ref}(P_{ref})
\end{align}
where $F_{ref}\in\mathbb{R}^{h\times w\times c}$ represents the extracted features. Since the input $P_{ref}$ is in 2D, we design the reflectance encoder $E_{ref}$ using 2D depthwise separable convolution blocks~\cite{sandler2018mobilenetv2} equipped with Instance Normalization~\cite{ulyanov2016instance} . Similar to the geometric encoder, spatial size and appearance context are progressively aggregated through a series of convolution blocks. Notably, the spherical projection is invertible, enabling the extracted reflectance features to be mapped back to their corresponding locations in 3D space.

We recognize that while recent works~\cite{xu2021rpvnet,alnaggar2021multi,li2022cpgnet} adopt multi-branch frameworks for point cloud segmentation, our approach is fundamentally different in three critical ways. First, we tackle the challenging domain generalization problem, which is ignored by~\cite{xu2021rpvnet,alnaggar2021multi,li2022cpgnet}. Second, we delve into the distinct domain shifts in geometry and reflectance under adverse weather, which are often treated as unified inputs in previous works. This insight leads to a specialized dual-branch encoding architecture to achieve better generalization ability. Moreover, we introduce a novel fusion mechanism specifically designed to handle corruption and distortion in point clouds due to adverse weather, an issue unaddressed in~\cite{xu2021rpvnet,alnaggar2021multi,li2022cpgnet}.

\subsection{Robust Multi-level Feature Collaboration}
\label{sec:method_umc}
After obtaining the geometric and reflectance features, the next step is to fuse them. However, simple fusion strategies, such as concatenation or addition, are often less effective for domain generalized LiDAR semantic segmentation, as the extracted features can still be impacted by noise and distortions from adverse weather conditions. To effectively harness useful information from both features while minimizing mutual interference, in this part,we propose a robust multi-level feature collaboration mechanism, which further augment features with complementary task-relevant information and then perform fusion at local and global levels.

\subsubsection{Complementarity-aware Information Constraint}
To encourage the encoders to extract discriminative and complementary information from the scene, we impose information constraints on the extracted geometric and reflectance features. To achieve this, we design a Complementarity-aware Information Constraint inspired by the recent success of variational information bottlenecks~\cite{peng2018variational, ahuja2021invariance, alemi2016deep}. Specifically, for a feature vector $f\in\mathcal{R}^c$ from a given feature map, we employ two fully connected networks to estimate the corresponding mean and variance vectors for each element,
\begin{align} 
\mu = \mathcal{W}_{\mu}(f), \quad \sigma = \mathcal{W}_{\sigma}(f) 
\label{eq:mean_var}
\end{align}
where $\mathcal{W}_{\mu}(\cdot)$ and $\mathcal{W}_{\sigma}(\cdot)$ are composed fully connected layers, and  $\{\mu\in\mathcal{R}^{c},\sigma\in\mathcal{R}^{c}_{+}\}$ are the mean and standard deviation vectors.
To ensure that $\sigma$ remains positive, $\mathcal{W}_{\sigma}(\cdot)$ is followed by a softplus activation function~\cite{dugas2000incorporating}. We then define the enhanced feature vector $m$ using reparameterization, which conforms to a multivariate Gaussian distribution characterized by the mean $\mu$ and standard deviation $\sigma$:
\begin{align} 
m = \mu + \epsilon \cdot \sigma, \quad \epsilon \sim \mathcal{N}(0, \mathbf{I}) 
\end{align}
where $\mathbf{I}$ is an identity matrix, and $\epsilon$ represents noise sampled from a standard Gaussian distribution. Then, we separately apply such distributional projection for feature vectors in the geometric feature map $F_{geo}$ and the reflectance feature map $F_{ref}$, as illustrated in Fig.~\ref{fig:architecture} (b). 

For each feature vector $f_{ref}$ in $F_{ref}$, we map its location to 3D volume and locate the corresponding feature vector $f_{geo}$ in $F_{geo}$. Then we have paired feature distributions, 
\begin{equation}
    \begin{split}
    p(m_{geo} \mid f_{geo}) &\sim \mathcal{N}(\mu_{geo},~\sigma_{geo}^2 \mathbf{I}) \\
    p(m_{ref} \mid f_{ref}) &\sim \mathcal{N}(\mu_{ref},~\sigma_{ref}^2 \mathbf{I})
\end{split}
\label{eq:dist}
\end{equation}
To explicitly enhance the robustness and complementarity of the geometric and reflectance features, we adopt the following objective:
\begin{equation}
\begin{split}
    \mathcal{L}_{cic} = &  \mathcal{KL}\left(p(m_{geo} \mid f_{geo}) \parallel \mathbf{r}(m_{geo})\right) \\
                      & + \mathcal{KL}\left(p(m_{ref} \mid f_{ref}) \parallel \mathbf{r}(m_{ref})\right) \\
                      & - \mathcal{KL}\left(p(m_{geo} \mid f_{geo}) \parallel p(m_{ref} \mid f_{ref})\right)\\
                      & - \mathcal{KL}\left(p(m_{ref} \mid f_{ref}) \parallel p(m_{geo} \mid f_{geo})\right)
\end{split}
\label{eq:cic}
\end{equation}
where $\mathcal{KL}(\cdot||\cdot)$ denotes the Kullback-Leibler divergence, and $\mathbf{r}(\cdot)$ represents a prior marginal distribution set as a standard Gaussian. Minimizing $\mathcal{L}_{cic}$ serves two primary purposes. First, it reduces the dependence between $m_{geo}$ and $f_{geo}$, as well as between $m_{ref}$ and $f_{ref}$, allowing $m_{geo}$ and $m_{ref}$ to discard domain-specific noise and capture more robust, generalizable, and task-relevant information. Second, it decreases the correlation between $m_{geo}$ and $m_{ref}$, thereby reducing redundancy and encouraging complementarity between the geometric and reflectance features. This dual effect ensures that the features are both robust and complementary, thereby enhancing the subsequent feature fusion and final prediction.

\subsubsection{Local-level Fusion}
So far, we have established distributional feature representations for each vector in the voxel-based 3D geometric feature map $F_{geo}$ and the range-view-based 2D reflectance feature map $F_{ref}$. To facilitate local fusion, we retrieve reflectance features for geometric feature vectors via spherical projection. 
Then, we perform local fusion by combining geometric feature with reflectance features.
Given the a geometric feature $\{\mu_{geo},\sigma_{geo}\}$ and the corresponding reflectance feature $\{\mu_{ ref},\sigma_{ ref}\}$, we apply a dynamic fusion approach, defined as:
\begin{equation}
    \begin{split}
        f_{local} = \alpha \cdot \mu_{geo} + (1 - \alpha) \cdot \mu_{ref}
    \end{split}
    \label{eq:local_fuse}
\end{equation}
where $\alpha = \frac{e^{\frac{1}{\Bar{\sigma}_{geo}}}}{e^{\frac{1}{\Bar{\sigma}_{geo}}} + e^{\frac{1}{\Bar{\sigma}_{ref}}}}$ with $\Bar{\sigma}_{geo}$ and $\Bar{\sigma}_{ref}$ as the standard deviation averaged along the channel dimmension. By incorporating the standard deviations into the fusion weights, the model dynamically assesses the reliability and contribution of each feature, enhancing the robustness of the fused features. Eq.~\ref{eq:local_fuse} is applied to all the geometric feature vectors, thus resulting in locally fused feature map $F_{local}$.

\subsubsection{Global-level Fusion}
The reflectance maps under adverse weather contain substantial noise at the local scale, yet they convey clear object semantics at the global level. Therefore, we adopt a global-level fusion module to further exploit this valuable information. An intuitive approach is to use Cross-Attention~\cite{vaswani2017attention} to directly aggregate global context from reflectance features into the geometric space. However, this straightforward solution can face challenges in computational efficiency and may remain vulnerable to overall distortions in the reflectance features caused by adverse weather.
To address these limitations, we introduce a two-stage cross-attention mechanism, as illustrated in Fig.~\ref{fig:architecture} (c). Given the reflectance feature map $M_{ref} \in \mathbb{R}^{h \times w \times c}$, we introduce a set of learnable global-query tokens $Q \in \mathbb{R}^{m \times c}$, with $m \ll hw$. The reflectance feature map $M_{ref}$ is first converted into Key and Value features, which are aggregated by the query tokens $Q$ to produce intermediate global features. These intermediate global features are then further aggregated by the geometric features $M_{geo}$:
\begin{align}
    F_{global} = \mathcal{CA}(M_{geo},~\mathcal{CA}(Q, M_{ref}))
\end{align}
where $\mathcal{CA}(\cdot, \cdot)$ represents the cross-attention operation, $M_{geo}$and $M_{ref}$ are the processed geometric features and relectance features, respectively. 
The learnable query $Q$ serves as an intermediary, integrating global information from the range-view features across different perspectives and effectively passing it to the geometric information in the voxel domain. It avoids the direct computation of attention between two large-scale feature sets in a single step, optimizing computational efficiency while ensuring meaningful cross-domain interaction. During the cross-attention process, the learnable $Q$ not only facilitates the fusion but also compresses and refines the range-view's global features, making the information more compact and relevant.

\subsection{Decoding and Training}
\label{sec:method_tai}
Since both the local-level and global-level fusion modules are designed to transfer information from the reflectance feature to the geometric feature, the resulting features $F_{local}$ and $F_{global}$ are both represented in the voxelized 3D space. We concatenate these features at each location and pass the combined representation to a decoder network to generate the final predictions.
During training, we optimize the model using the following loss function:
\begin{align}
    \mathcal{L}_{total} = \mathcal{L}_{CE} + \beta \mathcal{L}_{cic} 
\end{align}
where $\mathcal{L}_{CE}$ is the standard cross-entropy loss, and $\mathcal{L}_{cic}$ is the information constraint loss defined in Eq.~\ref{eq:cic}.

\renewcommand\arraystretch{1.1}
\setlength{\tabcolsep}{0.8mm}{
\begin{table*}[ht]
    \centering
    \begin{footnotesize}
    \begin{tabular}{l|ccccccccccccccccccc|cccc|c}
        \toprule
        Method & \rotatebox{90}{car} & \rotatebox{90}{bi.cle} & \rotatebox{90}{mt.cle} & \rotatebox{90}{truck} & \rotatebox{90}{oth-v.} & \rotatebox{90}{pers.} & \rotatebox{90}{bi.clst} & \rotatebox{90}{mt.clst} & \rotatebox{90}{road} & \rotatebox{90}{parki.} & \rotatebox{90}{sidew.} & \rotatebox{90}{oth-g.} & \rotatebox{90}{build.} & \rotatebox{90}{fence} & \rotatebox{90}{veget.} & \rotatebox{90}{trunk} & \rotatebox{90}{terra.} & \rotatebox{90}{pole} & \rotatebox{90}{traf.} & \rotatebox{90}{D-fog} & \rotatebox{90}{L-fog} & \rotatebox{90}{Rain} & \rotatebox{90}{Snow} & mIoU \\
        \hline
        Oracle & 89.4 & 42.1 & 0.0 & 59.9 & 61.2 & 69.6 & 39.0 & 0.0 & 82.2 & 21.5 & 58.2 & 45.6 & 86.1 & 63.6 & 80.2 & 52.0 & 77.6 & 50.1 & 61.7 & 51.9 & 54.6 & 57.9 & 53.7 & 54.7 \\
        \hline
        \hline
        \multicolumn{ 25 }{c}{SemanticKITTI$\rightarrow$SemanticSTF}\\
        \hline
        MinkNet\cite{choy20194d} & 55.9 & 0.0 & 0.2 & 1.9 & 10.9 & 10.3 & 6.0 & 0.0 & 61.2 & 10.9 & 32.0 & 0.0 & 67.9 & 41.6 & 49.8 & 27.9 & 40.8 & 29.6 & 17.5 & 29.5 & 26.0 & 28.4 & 21.4 & 24.4 \\
        PolarMix\cite{xiao2022polarmix} & 57.8 & 1.8 & 3.8 & 16.7 & 3.7 & 26.5 & 0.0 & 2.0 & 65.7 & 2.9 & 32.5 & 0.3 & 71.0 & 48.7 & 53.8 & 20.5 & 45.4 & 25.9 & 15.8 & 29.7 & 25.0 & 28.6 & 25.6 & 26.0 \\
        MMD\cite{li2018domain} & 63.6 & 0.0 & 2.6 & 0.1 & 11.4 & 28.1 & 0.0 & 0.0 & \underline{67.0} & 14.1 & 37.9 & 0.3 & 67.3 & 41.2 & 57.1 & 27.4 & 47.9 & 28.2 & 16.2 & 30.4 & 28.1 & 32.8 & 25.2 & 26.9 \\
        PCL\cite{yao2022pcl} & 65.9 & 0.0 & 0.0 & 17.7 & 0.4 & 8.4 & 0.0 & 0.0 & 59.6 & 12.0 & 35.0 & 1.6 & 74.0 & 47.5 & \underline{60.7} & 15.8 & \underline{48.9} & 26.1 & 27.5 & 28.9 & 27.6 & 30.1 & 24.6 & 26.4 \\
        PointDR\cite{xiao20233d} & 67.3 & 0.0 & 4.5 & 19.6 & 9.0 & 18.8 & 2.7 & 0.0 & 62.6 & 12.9 & \underline{38.1} & 0.6 & 73.3 & 43.8 & 56.4 & 32.2 & 45.7 & 28.7 & 27.4 & 31.3 & 29.7 & 31.9 & 26.2 & 28.6 \\
        UniMix\cite{zhao2024unimix} & 82.7 & \textbf{6.6} & 8.6 & 4.5 & 15.1 & 35.5 & \textbf{15.5} & 37.7 & 55.8 & 10.2 & 36.2 & 1.3 & 72.8 & 40.1 & 49.1 & 33.4 & 34.9 & 23.5 & 33.5 & 34.8 & 30.2 & 34.9 & 30.9 & 31.4 \\
        RDA\cite{park2024rethinking} & \textbf{86.1} & \underline{4.8} & \underline{13.8} & \underline{39.7} & \underline{26.6} & \textbf{55.4} & \underline{8.5} & \underline{50.4} & 63.7 & \underline{14.9} & 37.9 & \underline{5.5} & \underline{75.2} & \textbf{52.7} & 60.4 & \textbf{39.7} & 44.9 & \underline{30.1} & \underline{40.8} & \underline{36.0} & \underline{37.5} & \underline{37.6} & \underline{33.1} & \underline{39.5} \\
        Ours & \underline{85.6} & 2.5 & \textbf{16.7} & \textbf{47.6} & \textbf{33.0} & \underline{52.9} & 4.2 & \textbf{60.0} & \textbf{70.3} & \textbf{20.7} & \textbf{42.2} & \textbf{12.4} & \textbf{78.2} & \underline{51.4} & \textbf{63.2} & \underline{38.3} & \textbf{53.0} & \textbf{32.1} & \textbf{43.3} & \textbf{38.1} & \textbf{40.1} & \textbf{38.1} & \textbf{34.1} & \textbf{42.5}\\
        \hline
        \hline
        \multicolumn{ 25 }{c}{SynLiDAR$\rightarrow$SemanticSTF}\\
        \hline
        MinkNet\cite{choy20194d} & 27.1 & \textbf{3.0} & 0.6 & 15.8 & 0.1 & 25.2 & 1.8 & 5.6 & 23.9 & 0.3 & 14.6 & 0.6 & 36.3 & 19.9 & 37.9 & 17.9 & 41.8 & 9.5 & 2.3 & 19.9 & 17.2 & 17.2 & 11.9 & 15.0 \\
        PolarMix\cite{xiao2022polarmix} & 39.2 & 1.1 & 1.2 & 8.3 & 1.5 & 17.8 & 0.8 & 0.7 & 23.3 & 1.3 & 17.5 & 0.4 & 45.2 & 24.8 & 46.2 & 20.1 & \underline{38.7} & 7.6 & 1.9 & 16.1 & 15.5 & 19.2 & 15.6 & 15.7 \\
        MMD\cite{li2018domain} & 25.5 & 2.3 & 2.1 & 13.2 & 0.7 & 22.1 & 1.4 & 7.5 & 30.8 & 0.4 & 17.6 & 0.2 & 30.9 & 19.7 & 37.6 & 19.3 & \textbf{43.5} & 9.9 & 2.6 & 17.3 & 16.3 & 20.0 & 12.7 & 15.1 \\
        PCL\cite{yao2022pcl} & 30.9 & 0.8 & 1.4 & 10.0 & 0.4 & 23.3 & \textbf{4.0} & 7.9 & 28.5 & 1.3 & 17.7 & \underline{1.2} & 39.4 & 18.5 & 40.0 & 16.0 & 38.6 & 12.1 & 2.3 & 17.8 & 16.7 & 19.3 & 14.1 & 15.5 \\
        PointDR\cite{xiao20233d} & 37.8 & 2.5 & \underline{2.4} & \underline{23.6} & 0.1 & 26.3 & \underline{2.2} & 3.3 & 27.9 & \underline{7.7} & 17.5 & 0.5 & 47.6 & 25.3 & 45.7 & 21.0 & 37.5 & 17.9 & 5.5 & 19.5 & 19.91 & 21.1 & 16.9 & 18.5 \\
        UniMix\cite{zhao2024unimix} & \textbf{65.4} & 0.1 & \textbf{3.9} & 16.9 & \textbf{5.3} & \textbf{32.3} & 2.0 & \textbf{19.3} & \underline{52.1} & 5.0 & \textbf{27.3} & \textbf{3.0} & 49.4 & 20.3 & \textbf{58.5} & 22.7 & 23.2 & \textbf{26.9} & \textbf{10.4} & \textbf{24.3} & \underline{22.9} & \textbf{26.1} & \textbf{20.9} & \underline{23.4} \\
        RDA\cite{park2024rethinking} & 39.3 & \underline{2.9} & 0.9 & 19.4 & 0.8 & 27.7 & \underline{2.2} & 3.8 & 42.5 & \textbf{9.4} & 21.6 & 0.3 & \underline{51.9} & \underline{33.5} & 47.3 & \underline{23.1} & 33.3 & \underline{23.2} & \underline{6.8} & 19.0 & 21.2 & 23.1 & 17.3 & 20.5 \\
        Ours & \underline{49.5} & 2.0 & 2.3 & \textbf{24.4} & \underline{1.8} & \underline{31.8} & \underline{2.2} & \underline{9.0} & \textbf{62.3} & 4.6 & \underline{25.1} & 0.2 & \textbf{59.1} & \textbf{35.0} & \underline{55.7} & \textbf{24.1} & 32.8 & 18.8 & 6.1 & \underline{21.7} & \textbf{23.5} & \underline{24.9} & \underline{20.5} & \textbf{23.5}\\
        \bottomrule
    \end{tabular}
    \vspace{-0.3cm}
    \caption{Generalized segmentation performance with SemanticKITTI and SynLiDAR as the source and SemanticSTF as the target.}
    \label{tab:s-stf}  
    \vspace{-0.3cm}
  \end{footnotesize}
\end{table*}
}

\section{Experiments}
We evaluate the generalization ability of our method under adverse conditions with three settings: SemanticKITTI $\rightarrow$ SemanticSTF, SynLiDAR $\rightarrow$ SemanticSTF, and SemanticKITTI $\rightarrow$ SemanticKITTI-C. In these settings, the clear-weather real dataset SemanticKITTI~\cite{behley2019semantickitti} and synthetic dataset SynLiDAR~\cite{xiao2022transfer} serve as the source domains, while the adverse-condition target domains are represented by SemanticSTF~\cite{xiao20233d} and SemanticKITTI-C~\cite{kong2023robo3d}.

\subsection{Experimental details}
\textbf{Datasets.} 
\textit{SemanticKITTI}\cite{behley2019semantickitti} is a large-scale semantic segmentation dataset based on the KITTI Vision Benchmark\cite{geiger2012we}. The data is collected with a 64-beam LiDAR and annotated with over 19 semantic categories. Following the official protocol, we use sequences 00-07 and 09-10 as the training set, and sequence 08 as the validation set. 
\textit{SynLiDAR}\cite{xiao2022transfer} is a large-scale synthetic LiDAR dataset derived from various virtual environments, consisting of 13 sequences of LiDAR point clouds with approximately 20,000 scans. We split the training and validation sets in line with previous works\cite{xiao20233d, park2024rethinking, zhao2024unimix, xiao2022transfer}. 
\textit{SemanticSTF}\cite{xiao20233d} is an adverse-weather LiDAR segmentation dataset that extends the realistic STF Detection Benchmark\cite{bijelic2020seeing} by providing point-wise annotations for 21 semantic categories. It includes four common adverse weather conditions: dense fog, light fog, snow, and rain. We follow the official protocol and utilize its validation set, with 19 semantic categories, for testing.
\textit{SemanticKITTI-C}\cite{kong2023robo3d} is a corrupted LiDAR segmentation dataset created by applying eight types of corruption at three severity levels to the validation set of SemanticKITTI\cite{behley2019semantickitti}. We use mean Intersection over Union (mIoU) as the evaluation metric in our experiments.

\noindent\textbf{Implementation details}. 
Unless otherwise specified, we use the encoder and decoder of MiniNet-18/16~\cite{choy20194d} as our geometric encoder and task decoder, respectively. For the reflectance input, we employ an encoder built with the efficient Inverted Residual Blocks~\cite{sandler2018mobilenetv2} and Instance Normalization~\cite{ulyanov2016instance}. Fusion is performed on the outputs of the two encoders.
To train the network, we apply standard data augmentations, including random dropping, rotation, flipping, scaling, etc. We use the momentum SGD optimizer with a momentum of 0.9, a weight decay of 0.0001, and a batch size of 6 for both SemanticKITTI and SynLiDAR. The learning rate is initially set to 0.24 and is adjusted with the OneCycleLR policy~\cite{smith2019super} for 50 epochs in total. All experiments are conducted on a single Nvidia RTX4090 GPU.

\subsection{Main Results}

\textbf{SemanticKITTI to SemanticSTF/SemanticKITTI-C}. 
As shown in the upper part of Tab.~\ref{tab:s-stf}, our proposed method demonstrates significant improvements of the generalization performance on SemanticSTF dataset. Specifically, compared to the widely adopted baseline MinkNet, our approach achieves an impressive +18.1 increase in mIoU. Additionally, our method surpasses the state-of-the-art method RDA~\cite{park2024rethinking}, by +3.0 in mIoU.
This substantial performance gain is attributed to our strategy of decoupling geometric and appearance information, which effectively reduces negative impact of distortions and corruptions. Our method demonstrates the best results across all tested adverse weather conditions. Specifically, we observe mIoU improvements of +2.1 in dense fog, +2.6 in light fog, +0.5 in rain, and +1.0 in snow over the  state-of-the-art~\cite{park2024rethinking}. Compared with the MinkNet baseline, our method achieves remarkable increases of +8.6 mIoU in dense fog, +14.1 mIoU in light fog, +9.7 mIoU in rain, and +12.7 mIoU in snow.
Beyond performing well across adverse weather conditions, our method also yields superior results across most of the semantic categories. Compared to the baseline, we achieve more than +10 mIoU improvement in 13 categories. Notably, our approach shows gains of over +5 mIoU compared to the  state-of-the-art method~\cite{park2024rethinking} in categories such as trucks, other vehicles, motorcyclists, roads, parking, and terrain.
It is noteworthy that the previous state-of-the-art method~\cite{park2024rethinking} relies on reinforcement learning for data augmentation, requiring four A6000 GPUs for training. In contrast, our approach achieves superior performance through a simpler framework, without the need for such complex augmentation techniques, and can be trained efficiently on a single RTX 4090 GPU.

To further evaluate our method's effectiveness, we conduct experiments with more baseline network and datasets. As shown in Tab.~\ref{tab:s-kittic}, with SPVCNN~\cite{tang2020searching} or MinkNet18/32~\cite{choy20194d}, our method outperforms the previous state-of-the-art on both SemanticSTF and SemanticKITTI-C datasets with significant gaps. This again demonstrates the superioity of our proposed framework.

\setlength{\tabcolsep}{2.7mm}{
\begin{table}[t]
    \begin{footnotesize}
    \centering
    \begin{tabular}{l|cc}
    \toprule
    Method & SemanticSTF & SemanticKITTI-C \\
    \hline
    SPVCNN & 28.1 & 52.5 \\
    SPVCNN+RDA~\cite{park2024rethinking} & 38.4 & 52.9 \\
    SPVCNN+Ours & \textbf{40.8} &\textbf{54.2} \\
    \hline
    MinkNet18/32 & 31.4 & 53.0 \\
    MinkNet18/32+RDA~\cite{park2024rethinking} & 39.5 & 58.6 \\
    MinkNet18/32+Ours & \textbf{43.7} &\textbf{60.3} \\
    \bottomrule
    \end{tabular}  
    \vspace{-0.3cm}
    \caption{Generalized segmentation performance on SemanticSTF  and SemanticKITTI-C.}
    \vspace{-0.5cm}
    \label{tab:s-kittic}
    \end{footnotesize}
\end{table}}

\noindent\textbf{SynLiDAR to SemanticSTF}. 
The transition from SynLiDAR to SemanticSTF poses an even greater challenge than the previous setting, due to the substantial disparity between synthetic clear weather environments and real-world adverse weather conditions. As shown in the lower part of Tab.~\ref{tab:s-stf}, our method achieves the best overall performance in this challenging context, slightly outperforming the state-of-the-art UniMix method~\cite{zhao2024unimix}. While the improvement may appear modest, it is noteworthy that UniMix relies on complex weather simulations and exponential
moving average (EMA) ensembling~\cite{tarvainen2017mean}, which result in increased complexity and heavy resource consumption. In contrast, our method achieves competitive performance without requiring these intricate techniques, highlighting its efficiency and effectiveness in adapting to challenging environments. 


\subsection{Method Analysis}

\setlength{\tabcolsep}{1.8mm}{
\begin{table}[t]
    \begin{footnotesize}
    \centering
    \begin{tabular}{ccccc|cccc|c}
    \toprule
    GB &RB &CIC &LF  &GF &D-fog & L-fog & Rain & Snow & All\\
    \hline
                    &             &              &              &               & 29.5 & 26.0  & 28.4 & 21.4 & 24.4\\
    \hline
    $\checkmark$    &             &              &              &               & 32.3 & 37.7  & 35.4 & 32.1 & 37.3\\
    $\checkmark$    &$\checkmark$ &              &              &               & 31.6 & 35.7  & 35.1 & 32.8 & 36.5 \\
    $\checkmark$    &$\checkmark$ &$\checkmark$  &              &               & 33.1 & 40.5 & 36.8 & 33.4 & 38.6 \\
    $\checkmark$    &$\checkmark$ &$\checkmark$  &$\checkmark$  &               & 36.6 & 39.3  & 37.8 & 33.4 & 41.2\\
    $\checkmark$    &$\checkmark$ &$\checkmark$  &$\checkmark$  &$\checkmark$   & 38.1 & 40.1  & 38.1 & 34.1 & 42.5\\
    \bottomrule
    \end{tabular}
    \vspace{-5pt}
    \caption{Generalized segmentation performance on SemanticSTF  with different components. ``GB'' is the geometric encoder. ``RB'' is the reflectance encoder. ``CIC'' , ``LF'' and ``GF''denotes the complementarity-aware information constraint, local fusion, and global fusion, respectively. }
    \label{tab:ablation}
    \end{footnotesize}
      \vspace{-0.6cm}
\end{table}
}

\begin{figure*}[t]
    \centering
    \includegraphics[width=\linewidth]{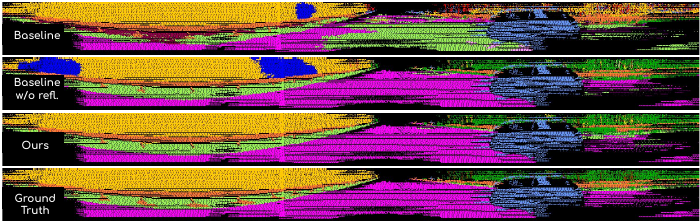}
      \vspace{-0.7cm}
    \caption{Qualitative results of our method on SemanticSTF. Without reflectance, the baseline achieves better overall segmentation accuracy but introduces new mispredictions, such as the blue region on the left half of the second image. Our method successfully corrects these mispredictions while maintaining overall accuracy.}
      \vspace{-0.5cm}
    \label{fig:quali}
\end{figure*}

\noindent\textbf{Ablation on Method Desgin.} We first analyze the effectiveness of the different designs of our method in Tab.~\ref{tab:ablation}.
The first row represents MinkNet, which is the baseline model and achieves the lowest accuracy. 
The second row indicate our geometric-encoder-only model, which is equivalent to the baseline without reflectance as input. By simply excluding reflection intensity, we observed a significant improvement of baseline across all adverse weather conditions, achieving a +12.9 mIoU overall.
This shows the severe negative impact of drastic changes in reflectance intensity distribution on robustness. 
In the third row, we introduce a reflectance encoder and simply merge the features by addition. However, we observe that the overall accuracy drops by 0.8 mIoU. This shows that the features are suffering from noise and distortion. Yet it is worth noting that even with minor degradation, the accuracy is still much higher than the baseline, which also demonstrate the  rationality of separating the feature extraction.
In the forth row, we apply the proposed information constraint loss during training and simply adding the processed features. We find that the performance gets improved  and outperforms the baseline model without reflectance, showing the  effectiveness of the propose information constraint.
In the fifth row, when introducing the Complementarity-aware Information Constraint with local fusion, the model achieves improvements in dense fog(+7.1 mIoU), light fog(+13.3 mIoU), rain(+9.4 mIoU), snow(+12.0 mIoU), and an overall performence increase of +16.8 mIoU. 
This shows the importance of the proposed information constraint, which alleviates noise interference and weather-relate distortion , and the stochastic modeling it introduces makes the model more robust to domain shifts. 
In the last row, further applying the global fusion results in improvements in dense fog(+8.6 mIoU), light fog(+14.1 mIoU), rain(+9.7 mIoU), snow(+12.7 mIoU), indicating that the success extraction of global information.
\begin{wrapfigure}{r}{0.2\textwidth}
    \centering
    \includegraphics[width=1\linewidth]{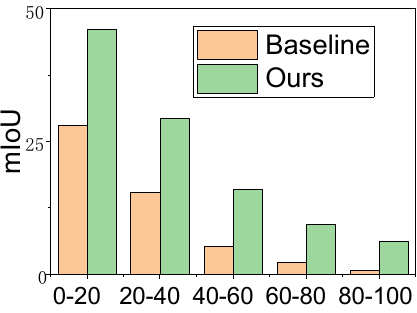}
      \vspace{-0.6cm}
    \caption{Generalized segmentation performance across different distance on SemanticSTF.}
      \vspace{-0.5cm}
    \label{fig:dist}
\end{wrapfigure} 
In addition, removing reflectance intensity reduces MinkNet's performance from 63.8 mIoU to 62.6 mIoU in the source domain, while GRC raises it to 63.1 mIoU, showing its effectiveness in utilizing both geometric and reflectance cues.
In Fig.~\ref{fig:dist}, we also show generalized segmentation performance across different distance intervals. As shown, our method achieve significant performance gains compared to the baseline model. We visualize some qualitative results in Fig.~\ref{fig:quali}
\setlength{\tabcolsep}{2.5mm}{
\begin{table}[t]
    \begin{footnotesize}
    \centering
    \begin{tabular}{ccc|c}
    \toprule
    Addition~\cite{alnaggar2021multi} &Gated Fusion~\cite{xu2021rpvnet} & Cross-Atten.~\cite{vaswani2017attention}  & Ours\\
    \hline
    36.5 &37.7&37.4&42.5\\
    \bottomrule
    \end{tabular}
      \vspace{-0.3cm}
    \caption{Impact of different fusion strategies. }
    \label{tab:fusion}
    \end{footnotesize}
      \vspace{-0.6cm}
\end{table}

\noindent\textbf{Fusion Strategy.}
In Tab.~\ref{tab:fusion}, we compare the impact of different fusion strategies that can be applied in our framework. As shown, fusing the geometric feature and reflectance features by simply addition leads to the worst performance, while applying gate fusion~\cite{xu2021rpvnet} or cross-attention~\cite{vaswani2017attention} improves the accuracy by about +1 in mIoU. 
We also observe that directly applying cross-attention between the two types of features incurs out-of-memory when trained with a RTX4090 GPU. In contrast, our proposed method achieves the best performance without exhausting the memory, demonstrating it efficacy.
We also experiment by progressively adding extra RMFC modules to shallower blocks in the encoders. However, the results in the target domain consistently showed a decline in performance. This may be caused by the fact that shallower features convey more heavily noise and distortion.

\setlength{\tabcolsep}{2.2mm}{
\begin{table}[t]
    \begin{footnotesize}
    \centering
    \begin{tabular}{l|ccc|c}
    \toprule
     & MACs (G) & Param (M) & FPS & mIoU \\
     \hline
    MinkNet18/32 & 141.7 & 21.7 & 19 & 31.4 \\
    MinkNet18/16 & 37.5 & 5.4 & 25 & 24.4 \\
    \hline
    MinkNet18/16+Ours & 70.8 & 8.4 & 19 & 42.5 \\
    \bottomrule
    \end{tabular}
      \vspace{-0.3cm}
    \caption{Analsyis of model size and inference complexity. }
    \label{tab:efficiency}
      \vspace{-0.6cm}
    \end{footnotesize}
\end{table}

\noindent\textbf{Efficiency  Analysis.}
\cref{tab:efficiency} presents the efficiency analysis of our proposed method alongside the baseline model. Compared to the baseline MinkNet18/16, our approach modestly increases model size and complexity, yet achieves a significant improvement in performance, with a +18$\%$ boost in mIoU. In contrast, MinkNet18/32 yields only a +7$\%$ mIoU increase despite its much larger model size and higher computational cost. This discrepancy highlights that the efficiency and effectiveness of our model stem from the inherent strengths of the proposed architectural design.

\section{Conclusion}
In this paper, we introduced a novel geometry-reflectance collaboration framework for generalized LiDAR semantic segmentation under adverse weather conditions. We delved into the challenge posed by heterogeneous domain shifts in the geometric structure and reflectance intensity of point clouds and proposed a solution that explicitly separates feature extraction for geometry and reflectance. Additionally, we developed a robust multi-level feature collaboration module to suppress redundant and unreliable information from both branches, achieving more robust and generalizable segmentation. Our results demonstrate that the propose approach outperforms prior methods, establishing new state-of-the-art performance.

\newpage

{
    \small
    \bibliographystyle{ieeenat_fullname}
    \bibliography{main}
}


\end{document}